\documentclass{article}
\usepackage{arxiv_package}

\usepackage[utf8]{inputenc} 
\usepackage[T1]{fontenc}    
\usepackage{hyperref}       
\usepackage{url}            
\usepackage{booktabs}       
\usepackage{amsfonts}       
\usepackage{nicefrac}       
\usepackage{microtype}      
\usepackage{xcolor}         

\usepackage{natbib}
\usepackage{caption, subcaption}
\usepackage{tikzit, tikzscale}

\tikzstyle{normal}=[fill=white, draw=black, shape=circle, minimum size=25pt, inner sep=0]
\tikzstyle{unobserved}=[fill={rgb,255: red,191; green,191; blue,191}, draw=black, shape=circle, minimum size=25pt, inner sep=0]
\tikzstyle{manipulable}=[fill={rgb,255: red,185; green,212; blue,255}, draw=black, shape=circle, minimum size=25pt, inner sep=0]
\tikzstyle{muct}=[fill={rgb,255: red,255; green,191; blue,191}, draw=black, shape=circle, inner sep=0, minimum size=25pt]
\tikzstyle{non manipulable}=[fill={rgb,255: red,246; green,194; blue,168}, draw=black, shape=circle, inner sep=0, minimum size=25pt]
\tikzstyle{factor}=[fill={rgb,255: red,207; green,219; blue,255}, draw=black, shape=rectangle, inner sep=0, minimum size=25pt]
\tikzstyle{long normal}=[fill=white, draw=black, shape=rounded rectangle, minimum height=17pt]
\tikzstyle{long factor}=[fill={rgb,255: red,204; green,255; blue,255}, draw=black, shape=rounded rectangle, minimum height=17pt]
\tikzstyle{Large factor}=[fill=none, draw=black, shape=rectangle, minimum height=30pt, minimum width=130pt, rounded corners]
\tikzstyle{new style 0}=[fill={rgb,255: red,204; green,255; blue,204}, draw=black, shape=rounded rectangle, minimum height=17pt]
\tikzstyle{new style 1}=[fill=none, draw=none, shape=circle, text width=10cm]

\tikzstyle{right}=[->, -latex, line width=0.25mm]
\tikzstyle{left}=[<-, latex-]
\tikzstyle{thin}=[-, line width=0.15mm]
\tikzstyle{thick}=[-, line width=0.25mm]
\tikzstyle{unobserved confounder}=[<->, dashed, latex-latex]
\tikzstyle{right added}=[->, -latex, draw=red, line width=0.25mm]
\tikzstyle{unobserved confounder added}=[<->, latex-latex, line width=0.25mm, draw=red, dashed]
\tikzstyle{right arrow blue}=[->, line width=0.25mm, draw={rgb,255: red,108; green,142; blue,191}, fill=none, -latex]
\tikzstyle{right arrow green}=[draw={rgb,255: red,24; green,172; blue,14}, ->, -latex, line width=0.25mm]

\usepackage{bbm}
\usepackage{multirow, arydshln, colortbl}
\usepackage{wrapfig}
\usepackage{amsfonts, amsmath}
\usepackage{algorithm2e}
\usepackage{cleveref}
\usepackage{textcomp}
\usepackage{pifont}
\usetikzlibrary{shapes,arrows,chains, positioning}
\usetikzlibrary[calc]
\usepackage{stackengine,collcell,makecell}

\newcolumntype{C}{>{\centering\arraybackslash}X}

\hypersetup{
    colorlinks=true,
    citecolor = blue,
    linkcolor=blue
}

\definecolor{tablenode}{HTML}{33FFFF}
\definecolor{lightcyan}{rgb}{0.88,1,1}
\definecolor{lightgray}{rgb}{.1, .1, .1}

\newcommand{\cmark}{\ding{51}\hfill}
\newcommand{\xmark}{\textcolor{black!20}{\ding{55}}\hfill}
\newcommand{\indep}{\perp \!\!\! \perp}

\title{Disentangled Counterfactual Recurrent Networks for Treatment Effect Inference over Time}

\author[1]{Jeroen Berrevoets\thanks{Corresponding author: \href{mailto:jeroen.berrevoets@maths.cam.ac.uk}{jeroen.berrevoets@maths.cam.ac.uk}}}
\author[1]{Alicia Curth}
\author[2,3]{Ioana Bica}
\author[1]{\\Eoin McKinney}
\author[1,2,4]{Mihaela van der Schaar}
\affil[1]{University of Cambridge}
\affil[2]{The Alan Turing Institute}
\affil[3]{University of Oxford}
\affil[4]{University of California, Los Angeles (UCLA)}

\begin{document}

\maketitle

\begin{abstract}
  Choosing the best treatment-plan for each individual patient requires accurate forecasts of their outcome trajectories as a function of the treatment, over time. 
  While large observational data sets constitute rich sources of information to learn from, they also contain biases as treatments are rarely assigned randomly in practice. To provide accurate and unbiased forecasts, we introduce the {\it Disentangled Counterfactual Recurrent Network} (DCRN), a novel sequence-to-sequence architecture that estimates treatment outcomes over time by learning representations of patient histories that are disentangled into three separate latent factors: a treatment factor, influencing {\it only} treatment selection; an outcome factor, influencing {\it only} the outcome; and a confounding factor, influencing {\it both}. With an architecture that is completely inspired by the causal structure of treatment influence over time, we advance forecast accuracy {\it and} disease understanding, as our architecture allows for practitioners to infer which patient features influence which part in a patient's trajectory, contrasting other approaches in this domain. We demonstrate that DCRN outperforms current state-of-the-art methods in forecasting treatment responses, on both real and simulated data.
\end{abstract}

\section{Introduction} \label{sec:introduction}

Clinical settings often require clinicians to monitor, and anticipate, the development of various biomarkers over time; for example in intensive care units \cite{johnson2016mimic}, or in patients with chronic diseases \cite{james2018global, butner2020mathematical}. In these settings, a clinician is tasked with influencing the future trajectory of biomarkers in a favorable direction, through curated treatment-plans. Identifying the optimal treatment-plan in a temporal setting is a complex undertaking as efficacy is influenced by longitudinal changes in natural history of disease and the extent of response to previous interventions \cite{huang2012analysis}. Treatment sequences are infrequently tested in clinical trials \cite{nolan2016use} and, where they are (for example through crossover trial design \cite{sibbald1998understanding}), sequences are short (typically limited to 2 stages), results are hampered by inevitable carryover of treatment effects between stages, and application is restricted to stable conditions with rapid on/offset of treatment effect \cite{nolan2016use}.

{\bf Forecasting treatment outcomes is important.} For example, Crohn’s disease is a relapsing, incurable autoinflammatory bowel disease for which {\it repeated courses of treatment are necessary} in the majority of patients \cite{neurath2017current}. Despite an increasing number of treatment options and clinical trials, it remains unclear how best to use available treatments in any { \it individual} patient \cite{noor2020personalised}. Hypertension is a prevalent, chronic condition in which treatment has unquestionably improved cardiovascular outcomes for many patients \cite{jones2020diagnosis}. However, with the majority of clinical trials pitting one treatment against another \cite{kaplan2007clinical}, treatment guidelines inevitably rely on expert clinical review of trial evidence to define the optimal sequence of treatments \cite{jones2020diagnosis} rather than test treatment sequences directly. There is therefore a large and unmet need to develop new and better methods that allow identification of optimal treatment sequences in a broad range of diseases. 

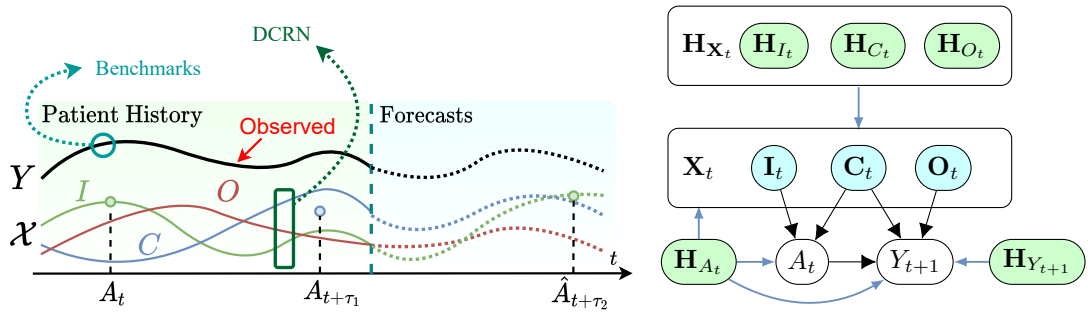
\begin{figure}[t]
    \centering
    \begin{subfigure}[b]{0.55\textwidth}
        \includegraphics[width=\textwidth]{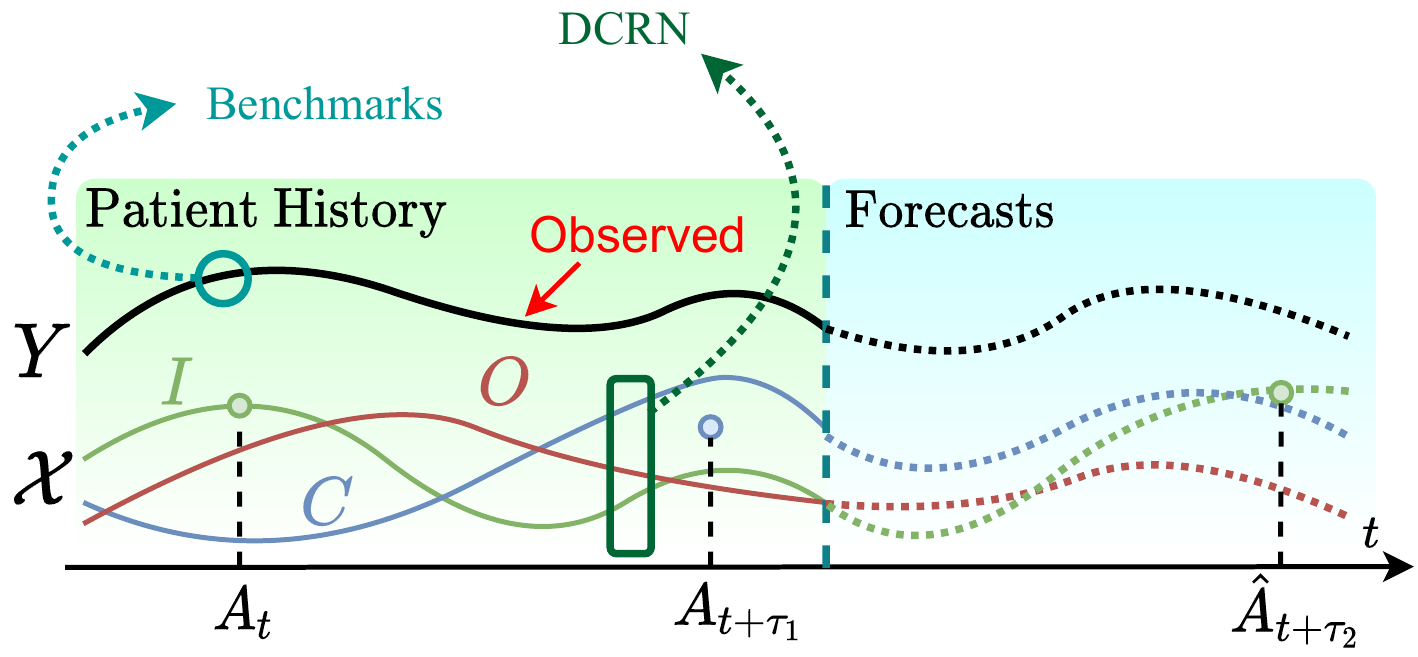}
    \end{subfigure}
    ~
    \begin{subfigure}[b]{0.35\textwidth}
        \begin{tikzpicture}
	\begin{pgfonlayer}{nodelayer}
		\node [style=new style 0] (0) at (2, -0.75) {$\mathbf{H}_{I_t}$};
		\node [style=new style 0] (1) at (4.5, -0.75) {$\mathbf{H}_{C_t}$};
		\node [style=new style 0] (2) at (7, -0.75) {$\mathbf{H}_{O_t}$};
		\node [style=long factor] (3) at (2, -4) {$\mathbf{I}_t$};
		\node [style=long factor] (4) at (4.25, -4) {$\mathbf{C}_t$};
		\node [style=long factor] (5) at (6.5, -4) {$\mathbf{O}_t$};
		\node [style=long normal] (6) at (2.75, -6.5) {$A_t$};
		\node [style=long normal] (7) at (5.75, -6.5) {$Y_{t+1}$};
		\node [style=new style 0] (8) at (9, -6.5) {$\mathbf{H}_{Y_{t+1}}$};
		\node [style=new style 0] (9) at (0, -6.5) {$\mathbf{H}_{A_t}$};
		\node [style=Large factor] (10) at (3.75, -4) {};
		\node [style=none] (11) at (0, -4) {$\mathbf{X}_t$};
		\node [style=Large factor] (12) at (3.75, -0.75) {};
		\node [style=none] (13) at (0.25, -0.75) {$\mathbf{H}_{\mathbf{X}_t}$};
		\node [style=none] (14) at (4.25, -1.85) {};
		\node [style=none] (15) at (4.25, -3) {};
		\node [style=none] (16) at (0, -5) {};
	\end{pgfonlayer}
	\begin{pgfonlayer}{edgelayer}
		\draw [->] (5) to (7);
		\draw [->] (4) to (7);
		\draw [->] (4) to (6);
		\draw [->] (6) to (7);
		\draw [style=right arrow blue] (8) to (7);
		\draw [->] (3) to (6);
		\draw [style=right arrow blue] (9) to (6);
		\draw [style=right arrow blue] (14.center) to (15.center);
		\draw [style=right arrow blue] (9) to (16.center);
		\draw [style=right arrow blue, bend right] (9) to (7);
	\end{pgfonlayer}
\end{tikzpicture}
    \end{subfigure}
    \caption{{\bf Overview of latent factors.} {\it Left} we have an illustration of the latent factors over time. The observed outcome trajectory ($Y$) is directly influenced by the "outcome factor" ($O$) and "confounding factor" ($C$); treatment selection is influenced by the "treatment factor" ($I$) and, again, the confounding factor. Rather than modeling $Y$ directly, DCRN models each factor--- and their various influences across time ---separately. {\it Right} we show our assumptions on the data-generating process (DGP) as a graphical model. We assume three latent factors $\mathbf{I}_t$, $\mathbf{C}_t$, and $\mathbf{O}_t$ influencing the action, the action \textit{and} the outcome, and the outcome, respectively; each influenced by their own history and past actions. }
    \label{fig:overview}
    \rule{\linewidth}{.75pt} 
\end{figure}

{\bf Why is this a hard problem?} While randomized experiments remain the gold standard for estimating treatment effects, there are many medical situations where randomization is either unethical, or simply infeasible. Due to the advent of electronic health records and other large patient databases, learning individualised treatment responses using \textit{observational} data is becoming a feasible option, \textit{if} biases in data can be accurately controlled for. Fortunately, recent years have seen a rapid increase in methods adapting the strengths of machine learning algorithms for causal inference from observational data (e.g. \cite{hill2011bayesian, johansson2016learning, shalit2017estimating, alaa2018limits, yoon2018ganite, hassanpour2020, zhang2020learning, berrevoets2020organite}), of which the main focus has been on the static setting.

The literature on counterfactual inference in the time series setting (e.g. \cite{schulam2017reliable, soleimani2017treatment, lim2018forecasting, bica2020estimating}), however, is much less mature than its static counterpart. In part, this is because the estimation of treatment effects over time is more difficult due to the presence of time-dependent confounders, which can themselves be affected by past administration of treatments \citep{robins2000marginal}. More specifically, inference in dynamic settings requires to identify not only how variables are related at a single point in time, but also a model of how such a dynamic system can be expected to evolve, leading to complete counterfactual \textit{trajectories} of arbitrary length. Such complexity warrants the need to develop more sophisticated model architectures.

\textbf{Contributions.} In this paper, we propose DCRN, a sequence-to-sequence architecture using recurrent neural networks, based on an assumed structural causal model of patient trajectories. We propose directly embedding the structure of a general temporal directed acyclic graph (DAG) into our neural architecture. Specifically, we propose to decompose the covariate space into three factors that have a direct effect only on treatment, on outcome, or both, as is illustrated in Figure~\ref{fig:overview}. This decomposition allows DCRN to better handle time-dependent confounding, resulting in a more accurate model. Next to accuracy, DCRN also has the potential to uncover interesting insights into underlying disease dynamics. As such, we make two key contributions:\\
 {\bf (1) Disentangled representations over time through a DAG-inspired network architecture.} 
 By directly modelling our proposed covariate space factorisation we learn a disentangled representation in the temporal setting and model how the various factors influence each other over time--- a major difficulty compared to the static setting. This is essential in medicine as the various covariates are part of a dynamic system that allows for factors to influence each other across time. Furthermore, with DCRN we gain insight into how the factored covariate space affects treatment assignment and outcome over time.\\
 {\bf (2) A more accurate sequence-to-sequence architecture for treatment effect estimation over time.}  Rather than learning a regression model directly on the covariate space--- unbiased either through weighting \cite{lim2018forecasting}, or domain adaptation \cite{bica2020estimating} ---DCRN uses only learned  {\it subsets} of the covariate space within its regression model and weighting mechanism. This allows balancing using a more targeted regularisation, avoiding unnecessary {\it over-regularisation}. We outperform existing benchmarks on real and simulated data.

\section{Related work}
We build on ideas from the literature on (i) representation learning for treatment effects in static settings and (ii) treatment effects over time, discussed in turn below.

\begin{table*}[t]
    \centering
    \caption{{\bf Overview of recurrent networks for temporal ITE estimation.} We summarise the most related benchmarks' architecture, key contribution, and merits. DCRN satisfies three key criteria: (1) account for selection bias; (2) no over-regularisation; (3) allow post-hoc network interpretation for analysis. DCRN achieves (2) and (3) through its DAG-inspired architecture.}
    \label{tab:benchmarks}
    
    \begin{tabularx}{\textwidth}{*{1}{l} C  l *{3}{c}}
    
        \toprule
        \textbf{Method}  & {\textbf{Overview}} & {\bf Contribution} & {\bf (1)} & {\bf (2)} & {\bf (3)} \\
        \toprule

        RMSN \citep{lim2018forecasting} &
        \begin{tikzpicture}[
            scale=.8,
            roundnode/.style={circle, draw=tablenode!60, fill=tablenode!10, very thick,  minimum size=4mm, inner sep=0},
            textnode/.style={rectangle, inner sep=0, minimum width=16mm}
        ]
            \node[roundnode] (data) at (0.5, 0)  {$\scriptstyle\mathcal{D}$};
            \node[roundnode] (prop) at (2.5, 0) {$\scriptstyle p_i$};
            \node[textnode] (outc) at (4, 0) {$\scriptstyle\min\{ p_i \mathcal{L}_Y\}$};
            \node[roundnode] (outcome) at (6, 0) {$\scriptstyle Y$};
            
            \draw[-latex] (data.east) -- node [left,above ] {$\scriptstyle p(A|X)$} (prop.west);
            \draw[-latex] (prop.east) -- (outc.west);
            \draw[-latex] (outc.east) -- (outcome.west);
            
            \draw[latex-] (prop.north west) to [out=135, in=90, looseness=1.3] (1.8, 0) to [out=270, in=225, looseness=1.3] (prop.south west);
        \end{tikzpicture}
        & \makecell[cl]{weighted\\ outcome loss} &\cmark& \cmark& \xmark  \\
        CRN \citep{bica2020estimating} &
        \begin{tikzpicture}[
            scale=.8,
            roundnode/.style={circle, draw=tablenode!60, fill=tablenode!10, very thick,  minimum size=4mm, inner sep=0},
            textnode/.style={rectangle, inner sep=0, minimum width=16mm}
        ]
            \node[roundnode] (data) at (0.5, 0) {$\scriptstyle \mathcal{D}$};
            \node[roundnode] (phi) at (1.7, 0) {$\scriptstyle \Phi$};
            \node[textnode] (prop) at (4.1, .2) {$\scriptstyle \max\{ \mathcal{L}_A \}$};
            \node[textnode] (est_loss) at (4.1, -.2) {$\scriptstyle \min\{\mathcal{L}_Y\}$};
            \node[roundnode] (outcome) at (6, -.2) {$\scriptstyle Y$};
            
            \draw[-latex] (data.east) -- (phi.west);
            \draw[-latex, rounded corners] (phi.east) -- (2.2, 0) -- (2.7, .2) -- (prop.west);
            \draw[-latex, rounded corners] (phi.east) -- (2.2, 0) -- (2.7, -.2) -- (est_loss.west);
            \draw[-latex] (est_loss.east) -- (outcome.west);
            
            \draw[latex-] (phi.north west) to [out=135, in=90, looseness=1.3] (1, 0) to [out=270, in=225, looseness=1.3] (phi.south west);

        \end{tikzpicture}
        & \makecell[cl]{balanced\\ representations} &\cmark& \xmark& \xmark  \\
        \midrule
        DCRN (ours) &
        \begin{tikzpicture}[
            scale=.8,
            roundnode/.style={circle, draw=tablenode!60, fill=tablenode!10, very thick,  minimum size=4mm, inner sep=0},
            textnode/.style={rectangle, inner sep=0, minimum width=5mm}
        ]

            \node[roundnode] (data) at (.5, 0) {$\scriptstyle\mathcal{D}$};
            \node[roundnode] (hist) at (1.5,0) {$\scriptstyle \mathbf{H}$};
            \node[roundnode] (phi) at (2.5, 0) {$\scriptstyle \Phi$};
            
            \node[textnode] (inst) at (4.1, .3) {$\scriptstyle I$};
            \node[textnode] (conf) at (4.1, 0) {$\scriptstyle C$};
            \node[textnode] (outc) at (4.1, -.3) {$\scriptstyle O$};
            
            \node[roundnode] (prop) at (6, .3) {$\scriptstyle A$};
            \node[roundnode] (outcome) at (6, -.3) {$\scriptstyle Y$};
            
            \draw[-latex] (data.east) -- (hist.west);
            \draw[-latex] (hist.east) -- (phi.west);
            
            \draw[-latex, rounded corners] (phi.east) -- (3, 0) -- (3.3, .3) -- (inst.west);
            \draw[-latex, rounded corners] (phi.east) -- (conf.west);
            \draw[-latex, rounded corners] (phi.east) -- (3, 0) -- (3.3, -.3) -- (outc.west);
            
            \draw[-latex] (inst.east) -- (prop);
            \draw[-latex, rounded corners] (conf.east) -- (4.5, 0) -- (5, .3) -- (prop.west);
            \draw[-latex, rounded corners] (conf.east) -- (4.5, 0) -- (5, -.3) -- (outcome.west);
            \draw[-latex] (outc.east) -- (outcome);
            
            \draw[latex-] (hist.north west) to [out=135, in=90, looseness=1.3] (.8, 0) to [out=270, in=225, looseness=1.3] (hist.south west);
        \end{tikzpicture}
        & \makecell[cl]{ disentangled\\ representations}& \cmark& \cmark& \cmark\\
        
        \bottomrule

    \end{tabularx}  
    
\end{table*}

\textbf{Representation learning for treatment effect estimation in the static setting.} Estimating heterogeneous treatment effects in the static setting  has received considerable attention in recent years. A wide range of machine learning methods have been adapted to do so, and, next to representation learning,  popular examples include tree-based methods (e.g. \cite{wager2018estimation, hill2011bayesian}), GANs \citep{yoon2018ganite} and Gaussian Processes \citep{alaa2018limits}. Our work follows the recent stream of literature relying on learning shared feature representations for treatment and control groups.  \cite{johansson2016learning} and \cite{shalit2017estimating} proposed to regularize towards finding \textit{balanced} representations, considering the main challenge a \textit{covariate shift} problem between treatment and control groups. Subsequently, this was extended to enforce the preservation of local similarity  \citep{yao2018representation}, to learn overlapping representations \citep{zhang2020learning} and to incorporate importance weighting  \citep{hassanpour2019counterfactual}. Instead of learning one representation of all inputs, \cite{hassanpour2020} and \cite{wu2020learning}  identify \textit{disentangled} representations that separate input covariates by the effects they have on treatment assignment and outcome. However, none of these methods are fit for use in dynamic settings, as they model neither time-dependent confounding nor other dynamic relationships between variables.

\textbf{Estimating treatment effects over time.} Traditional approaches to estimating treatment effects over time originate in the epidemiology literature, and are based on g-computation, Structural Nested Models and Marginal Structural Models (MSMs) \citep{robins1994correcting, robins2000marginal, robins2008estimation}. Because such models are usually highly parametrized and rely on linear predictors, they do not allow to consider complex non-linear temporal relationships. For this reason, recent literature has focussed on adapting Bayesian nonparametric approaches and recurrent neural networks to allow more flexible estimation. Bayesian nonparametric estimation of treatment effects in dynamic settings has relied mainly on Gaussian Processes in discrete time settings \citep{xu2016bayesian},  and continuous time settings \citep{soleimani2017treatment, schulam2017reliable}, or been used in combination with dirichlet processes \citep{roy2016bayesian}. Nonetheless, these methods make strong assumptions on the underlying structure of the problem, and do not model treatment effect heterogeneities flexibly.

\textbf{Individual treatment effects over time, using deep neural networks.} Our work is related most closely to \cite{lim2018forecasting} and \cite{bica2020estimating}, who use recurrent neural networks to model both time dependencies and treatment effect heterogeneities without strong assumptions on functional forms. \cite{lim2018forecasting} propose Recurrent Marginal Structural Networks (RMSNs), a sequence-to-sequence architecture, that--- inspired by MSMs ---uses inverse probability of treatment weighting to handle the bias from the time-dependent confounders. \cite{bica2020estimating}, on the other hand, propose the counterfactual recurrent network (CRN) to estimate outcomes by learning representations of the patient history that are invariant to the treatment assigned at each timestep. We propose to learn representations that are used for both regression adjustment and propensity weighting, as such combining the strengths of both methods within our model. We provide an overview of these benchmarks in Table~\ref{tab:benchmarks} where we focus on the differences between DCRN, CRN, and RMSN--- DCRN's architecture is discussed in more detail in Section~\ref{sec:dcrn}.

\section{Problem setting.} \label{sec:setup}
Assume that we observe a tuple $(\mathbf{X}, A, Y)^{(i)}_t = (\mathbf{X}_t^{(i)} , A^{(i)}_t, Y_t^{(i)}) \in \mathcal{X} \times \mathcal{A} \times \mathcal{Y} $ for each patient  $i \in \{1, \ldots, N\}$ for time steps $t = \{1, \ldots, T^{(i)}\}$. Here, $\mathbf{X}_t^{(i)} \in \mathcal{X} \subset \mathbb{R}^d$ denotes the $d$-dimensional vector of (time-varying) observed characteristics associated with a patient; $A^{(i)}_t \in \mathcal{A} = \{0, 1\}$ the administered treatment; and $Y_t^{(i)} \in \mathcal{Y} \subset \mathbbm{R}$ the clinical outcome. For notational convenience we will drop the index $i$ in the discussion below. 

In Figure \ref{fig:overview} (right) we illustrate our assumed data-generating process. In particular, we assume that the time-varying covariates $\mathbf{X}_t$ can be \textit{decomposed} into three components, so that we have the factors $(\mathbf{I}_t, \mathbf{C}_t, \mathbf{O}_t) = (\mathbf{I}_t(\mathbf{X}_t), \mathbf{C}_t(\mathbf{X}_t), \mathbf{O}_t(\mathbf{X}_t))$. Here, $\mathbf{I}_t$ affects only treatment assignment $A_t$; $\mathbf{O}_t$ only affects the next period outcome $Y_{t+1}$; so that $\mathbf{C}_t$ is the only true \textit{confounder} affecting both treatment assignment \textit{and} next period outcome. We further assume that the decomposed representation of $\mathbf{X}_t$ is \textit{unknown} and has to be learned. For any observation of any variable $V_t$ we denote by $\mathbf{H}_{V_t}=(V_{t-1}, V_{t-2},\ldots)$ its history. Note that, while we assume the possibility to disentangle, DCRN has a flexible architecture which will perform well even if the covariates in the underlying DGP cannot be disentangled, as all covariates will then fall within the confounding factor. We test DCRN for this situation in our synthetic setup, by letting treatment-selection be determined entirely by the confounding factor (Section~\ref{sec:experiments:synth}).

\textbf{Goal.} Within this setup, our goal is to build a model that accurately forecasts the $\tau$-step ahead sequence of expected potential outcomes given as input an entire treatment sequence $A_{t:t+\tau-1}=\tilde{a}_{t:t+\tau-1}$,
\begin{equation}
\{\mathbbm{E}[Y(\tilde{a}_{t:t+\tau-1})_{t+\tau}|\mathbf{H}_t]\}_{\tau \geq 1} \quad=\quad
\{\mathbbm{E}[Y_{t+\tau}|A_{t:t+\tau-1}=\tilde{a}_{t:t+\tau-1}, \mathbf{H}_t]\}_{\tau \geq 1}
\end{equation}   
for fixed patient history $\mathbf{H}_t=[X_t, \mathbf{H}_{\mathbf{X}_t}, Y_t, \mathbf{H}_{Y_t}, \mathbf{H}_{A_t}]$ (which includes all patient information up to time $t$ \textit{except} the treatment assignment $A_t$).  Here, we use the term \textit{forecasting} instead of \textit{predicting} in line with \cite{gische2020forecasting}, to signify that we aim to build a model which allows to give \textit{causal} interpretation to the modeled hypothetical effects of interventions. 


\textbf{Assumptions. } 
Because we are interested in the effects of interventions in dynamic settings, we consider our problem within the framework of structural causal models which were derived in econometrics to model endogenous responses in dynamical systems \citep{lewbel2019identification}. In Figure \ref{fig:overview} we note the graphical model with identifying restrictions that we assume in this paper. In structural models, identification of instantaneous effects is often achieved by enforcing exclusion restrictions through variable ordering \citep{lewbel2019identification}. Here, we place the restrictions that at any time step $t$, the covariates $X_t$ can affect treatment assignment $A_t$ (but not the reverse), and that both $X_t$ and $A_t$ affect only future outcomes $\{Y_{t+\tau}\}_{\tau\geq 1}$ (i.e. there are no instantaneous effects on same-period outcomes $Y_t$; that is a treatment in period $t$ would show effect at the earliest in period $t+1$ -- e.g. at the next doctor's appointment). In addition, analogous to the static causal inference setting, interpreting estimated dynamic effects as causal requires the imposition of assumptions that are untestable and have to be based on domain knowledge. In particular, we work under the standard assumption in sequential individual treatment effect estimation:
\begin{assumption}[No unmeasured confounding, Consistency, and Overlap through time] There are no unobserved variables affecting both treatment selection, and outcome: $A_t \indep Y_{t+1}|\mathbf{X}_t^{(i)}$ for all $i \in \{1,\dots,N\}$, and all $t\in\{1,\dots,T^{(i)}\}$ (illustrated in Figure~\ref{fig:overview}). If $A_t=a$ then $Y_{t+1} = Y_{t+1}(a)$. There is a non-zero probability for receiving treatment through time, $0< \mathbbm{P}(A_t=a | \mathbf{H}_t) < 1$ for all $t$ and $\mathbf{H}_t$ and $a$.
\end{assumption}
Note that, while the above assumptions are strong, they are standard in both the static \citep{shalit2017estimating, alaa2018limits} and sequential \citep{bica2020estimating, lim2018forecasting} individual treatment effects literature.
 
\section{DCRN} \label{sec:dcrn}
\begin{figure*}[t]
    \centering
    \includegraphics[width=0.8\textwidth]{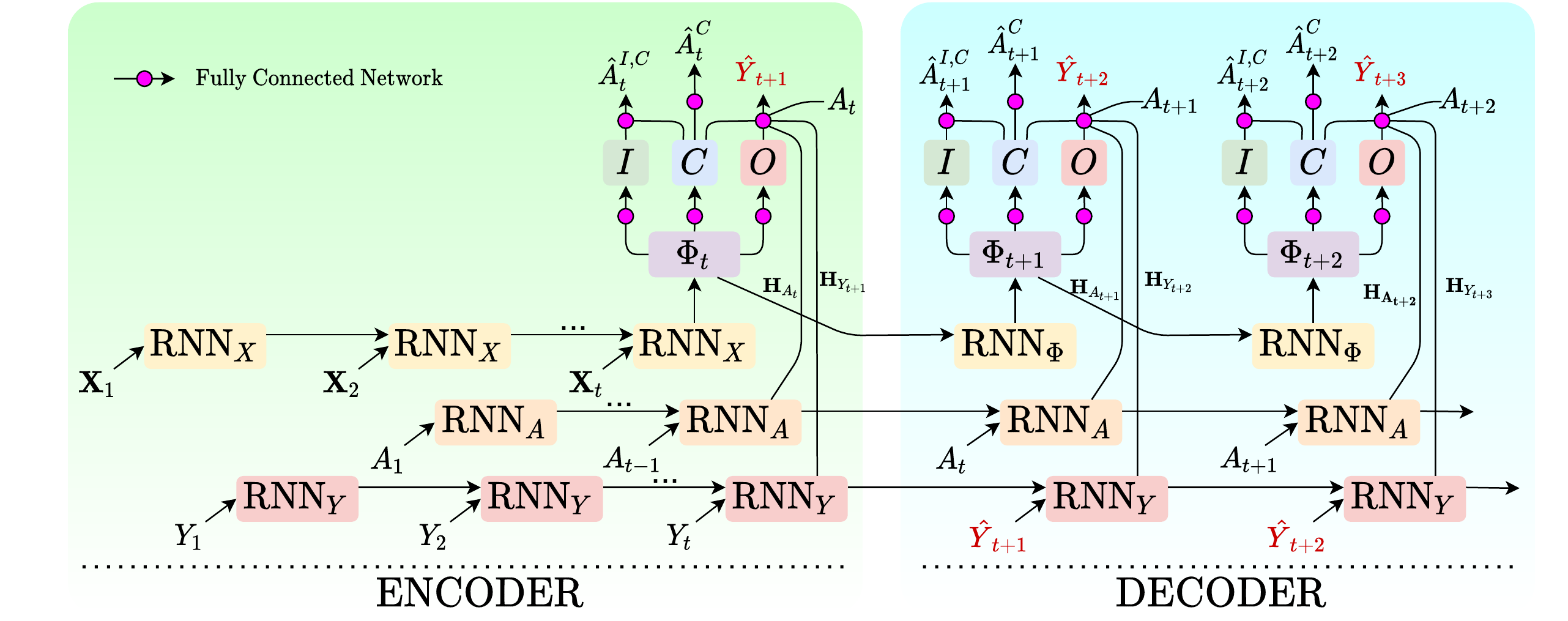}
    \caption{{\bf DCRN architecture.} We learn an encoder from the patient history, which jointly optimizes for representations and 1-step ahead forecasting. From the encoder, we then learn a decoder which autoregressively completes the patient's outcome and treatment trajectory.}
    \label{fig:DCRN:architecture}
    
\end{figure*}
We first  give an overview of our proposed method, the Distentangled Counterfactual Recurrent Network (DCRN), discussing the architecture, the encoder and the decoder in turn, and then discuss underlying loss functions.
\subsection{Overview of the method}
We propose DCRN, which learns disentangled representations of patient covariates from observational data, allowing to combine outcome weighting and representation learning during training, with the ultimate goal of performing counterfactual inference over time. To implement this in the dynamic setting, an obvious option would be to let the disentangled latent factors be represented by three distinct recurrent networks (RNNs), a diagram of this architecture can be found in our supplemental materials. Specifically, one could jointly optimise three distinct networks for $\mathbf{I}(\mathbf{X}_t)$, $\mathbf{C}(\mathbf{X}_t)$, and $\mathbf{O}(\mathbf{X}_t)$, used for weighting and adjustment in forecasting, respectively. While this is a viable approach, it is also a very restrictive one as it enforces that the disentangled factors are not able to influence one another over time. The dynamic model that we assume (Figure~\ref{fig:overview}) is much more general than this, also allowing for dynamic interaction between factors over time. To let one factor be dependent on another factor from a previous time-step, DCRN builds a joint representation of all covariates $\mathbf{X}_t$ using one RNN, which is then disentangled per time-step using fully-connected networks.


\textbf{DCRN Architecture.} DCRN consists of multiple recurrent and fully connected networks, as shown in Figure~\ref{fig:DCRN:architecture}. With these networks, DCRN follows our proposed graphical model in Figure~\ref{fig:overview}, where the assumed data-generating process (DGP) is informing which inputs enter which part of our architecture. In principle, our architecture could also be used when a domain expert considers a different underlying DGP more likely, which would only require changing inputs to different components of the model. To our knowledge, DCRN is the first temporal architecture in individualized treatment effect estimation designed to follow a DAG capturing the assumed interactions between variables in the underlying DGP.

Here, all history nodes (i.e. $\mathbf{H}_{\mathbf{X}_t}$, $\mathbf{H}_{Y_{t+1}}$, and $\mathbf{H}_{A_t}$) are represented by RNNs, each providing a representation of their given history. The representations of $\mathbf{H}_{Y_{t+1}}$ and $\mathbf{H}_{A_t}$ are used directly in the predictor for $\hat{Y}_{t+1}$--- thus implying that past treatments and outcomes still affect current outcomes ---while the representation for $\mathbf{X}_{t}$ (which we denote by $\Phi_t$) is first disentangled before it is used to forecast future outcomes $\hat{Y}_{t+1}$ and treatment assignments $\hat{A}_t$.  We have organised our architecture in an encoder-decoder structure, which enables making multiple step-ahead forecasts, as we discuss below.

\textbf{Encoding temporal covariates.} From the history of time-varying covariates, $\mathbf{H}_{\mathbf{X}_t}$, we learn a representation $\Phi_t$ which can subsequently be used to predict both the treatment, $A_t$, and the next outcome, $Y_{t+1}$. For this, we take $\Phi_t$ as input to: (i) a network whose sole task is to learn a representation $\mathbf{I}(\Phi_t)$, only able to predict the next action -- i.e. the treatment factor; (ii) a network tasked with learning a representation $\mathbf{O}(\Phi_t)$, only able to predict the next outcome -- i.e. the outcome factor; (iii) and a network that learns a representation $\mathbf{C}(\Phi_t)$ that can predict both -- i.e. the confounding factor. At each time-step there are three output-heads, which are used to predict: (i) the probability of treatment assignment $\hat{A}^{I,C}_{t}=\hat{\mathbbm{P}}(A=1|\mathbf{I}_t, \mathbf{C}_t)$; (ii) the probability of treatment assignment using only the confounding factor, $\hat{A}^C_{t} = \hat{\mathbbm{P}}(A=1|\mathbf{C}_t)$; and the next outcome, $\hat{Y}_{t+1} = \hat{\mathbbm{E}}(Y_{t+1}|\mathbf{C}_t, \mathbf{O}_t, \mathbf{H}_{A_t}, \mathbf{H}_{Y_{t+1}}, A_t)$. Note that $A_t$ is predicted twice as (i) is used to identify the treatment factor, but we weight by (ii) as only true confounders matter for balancing \cite{hassanpour2020}. This is illustrated in the leftmost part of Figure~\ref{fig:DCRN:architecture}. To enforce hard disentanglement where any dimension in $\mathcal{X}$ can only influence one latent factor, we employ a regularisation component in our loss, discussed below.

\textbf{Decoding the recurrent representation for patient trajectories.} The encoder is only able to predict the outcome one step ahead, because it requires as input $\mathbf{X}_{t-\tau}$ for each time step in the history, $\mathbf{H}_{\mathbf{X}_t}$. Using the encoder to predict multiple step ahead trajectories would thus require predicting the entire trajectory of patient covariates, which in large dimensions is infeasible. Instead, we compose a decoder that takes the latent representation, $\Phi_t$, of the encoder as input, and learns to autoregressively forecast the latent representation directly\footnote{When these encodings are estimated, they are detached from the computational graph such that a backpropagating gradient does no longer influence the encoder, and consequentially the one step-ahead forecasts.}. During training, this is ensured by optimising for a good fit on the same three output heads as in the encoder, with the exception that the output head also receives the previous outcome as input. At test time, the outcome is replaced with the predicted outcome of the previous time-step, depicted in the rightmost part of Figure~\ref{fig:DCRN:architecture}.

While our decoder is similar to the decoder used in \cite{bica2020estimating}, it varies in three major aspects: (i) as we are not balancing the representation by means of \textit{adversarial} training, i.e. \textit{maximizing} a prediction loss on $\hat{A}_t$, we learn also a valid propensity model, and could use treatment predictions based on $\hat{A}_t$ as an input to the decoder and perform impulse response modelling, a detailed explanation on this is provided in the supplemental material; (ii) we use the encoder's last output directly as input to the decoder, providing higher dimensional input used for disentangling in the decoder, contrasting CRNs \citep{bica2020estimating}; and (iii), the RNNs representing $\mathbf{H}_{Y_t}$ and $\mathbf{H}_{A_t}$ are continued in the decoding phase, albeit autoregressively in case of the outcomes.

\subsection{Loss function}
We translate our learning objectives into four major components in our loss function. To introduce these components, let $\theta_{\mathbf{H}}$ be the parameters for the history networks ($\mathbf{H}_{Y_t}$ and $\mathbf{H}_{A_t}$); $\theta_\Phi$ the parameters for the history network for $\mathbf{H}_{\mathbf{X}_t}$; $\theta_I$, $\theta_C$, and $\theta_O$ the parameters for the treatment, confounding, and outcome factor, respectively; $\theta_A$ the parameters for the treatment prediction; and $\theta_Y$ the parameters for the eventual forecast. Note that each component to our loss is linearly combined, where each coefficient corresponds to a hyperparameter. These hyperparameters are tuned using factual data (as we have no access to counterfactual data), and reported in our supplemental material.

The first component is a weighted mean squared error loss on the factual outcomes,
\begin{equation}
    \mathcal{L}_Y(\theta_{\mathbf{H}}, \theta_\Phi, \theta_C, \theta_O, \theta_Y) = \frac{1}{N} \sum_{i=1}^N \Big( \omega(A_t, \mathbf{C}_t) \cdot \lVert Y_{t+1}, y\big(\mathbf{C}_t, \mathbf{O}_t, \mathbf{H}_{Y_{t+1}}, \mathbf{H}_{A_t}, A_t\big)\rVert_2^2 \Big), \label{eq:loss:prediction}
\end{equation}
where the weighting function is defined, $\omega(A_t, \mathbf{C}_t) = 1 + \frac{\mathbbm{P}(\mathbf{C}_t | \neg A_t)}{\mathbbm{P}(\mathbf{C}_t | A_t)} = 1 + \frac{\mathbbm{P}(A_t)}{1 - \mathbbm{P}(A_t)}\cdot\frac{1-\hat{A}^C_t}{\hat{A}_t^C}$ as in \cite{hassanpour2020}, where the weights only depend on the true confounders ($\mathbf{C}_t$). Note that the instrumental factor, $\mathbf{I}_t$, has no influence on $\mathcal{L}_Y$, requiring $\mathbf{C}_t$ and $\mathbf{O}_t$ to retain information such that $\mathcal{L}_Y$ may be minimised. The second part of the loss function is a discrepancy measure that aims to balance the outcome factor,
\begin{equation}
    \mathcal{L}_D(\theta_\Phi, \theta_O) = \alpha \cdot \text{\texttt{disc}}\big( \{\mathbf{O}_t\}_{A_t=0}, \{\mathbf{O}_t)\}_{A_t=1} \big), \label{eq:loss:discrepancy}
\end{equation}
where $\alpha \in \mathbb{R}_+$ is the first hyperparameter of our model, governing the influence the discrepancy measure has on training. While we use a maximum mean discrepancy (MMD) as the discrepancy measure \citep{gretton2012kernel}, other discrepancy methods are also viable \citep{shalit2017estimating, hassanpour2020}. With $\mathcal{L}_D$ we force the representation, $\mathbf{O}_t$, to minimise the difference between the treated and controlled subsets of the data, making $\mathbf{O}_t$ unable to predict the treatment. Note that none of the confounders $\mathbf{C}_t$ are discarded by minimising $\mathcal{L}_D$. This is a major difference with \citet{bica2020estimating} where non-confounding covariates (i.e. those in the instrumental or outcome factor), are also "balanced", resulting in over-regularisation. The third component of our loss function is a cross entropy loss for both treatment prediction heads,
\begin{equation}
    \mathcal{L}_C(\theta_\Phi, \theta_I, \theta_C, \theta_A) = \beta \cdot \frac{1}{N}\sum_{i=1}^N \big( -\log(\hat{A}_t^{I,C})-\log( \hat{A}_t^{C}) \big), \label{eq:loss:propensity}
\end{equation}
with $\beta \in \mathbb{R}_+$ a second hyperparameter, controlling the influence of the treatment prediction heads. In effect, $\mathcal{L}_C$ achieves the opposite of what $\mathcal{L}_D$ contributes to the learned representation, that is it will make differences between treatment groups more apparent in its representation.

Finally, because we learn the dynamic model and decompose $\mathbf{X}_t$ into its factors using representation learning \textit{jointly}, we need to impose identifying restrictions also on the disentanglement to ensure that it can be uniquely identified. We impose that the disentangled factors do not rely on overlapping inputs\footnote{To see why this is necessary in a simple case, note that if $Y_{t+1}$ was linear in $A_t$, $\mathbf{C}_t$ and $\mathbf{O}_t$ only, and all factors were influenced by a joint set of covariates $\Phi^{ICO}$ we would have that $\mathbbm{E}[Y_{t+1}|A_t, \mathbf{X}_t] = \beta_A A_t + \beta_C \mathbf{C}_t + \beta_O \mathbf{O}_t$ and $\mathbbm{E}[Y_{t+1}|A_t, \mathbf{X}_t]= \beta_A A_t + \beta_C (\mathbf{C}_t+\beta_O f(\Phi^{ICO})) + \beta_O (\mathbf{O}_t - \beta_C f(\Phi^{ICO}))$ would lead to the same fit for any function $f$ and hence result in the same value of a loss function. This phenomenon is not unique to the dynamic case, and is also a problem in \cite{hassanpour2020}'s static model.}, i.e. ($\mathbf{I}_t(\Phi_t), \mathbf{C}_t(\Phi_t), \mathbf{O}_t(\Phi_t)$) = ($\mathbf{I}_t(\Phi^{I}_t), \mathbf{C}_t(\Phi^C_t), \mathbf{O}_t(\Phi^O_t)$) with $\Phi^J_t \cap \Phi^L_t = \emptyset \text{ for } J \neq L, \text{ and } J, L \in \{I, C, O\} $. \cite{wu2020learning} improve over \cite{hassanpour2020}'s method by augmenting their loss function to enforce that the factors do not overlap, an approach that we adapt to the dynamic setting. Note that forcing such disentanglement is not more restrictive than assuming complete entanglement -- should there be no underlying disentangled factors in the data-generating process, then DCRN will simply not learn these factors and will represent the entire covariate space through the confounding factor, and hence able to handle any violations against the assumed existence of disentangled factors.

Thus, while the above loss components will suffice to achieve \textit{some} disentanglement, we adapt \cite{wu2020learning}'s  orthogonal regularizer to enforce better identification of the factors, and add to our loss function,
\begin{equation}
    \mathcal{L}_O(\theta_I, \theta_C, \theta_O) = \gamma \big( \bar{W}_{\Phi:I} \times \bar{W}_{\Phi:C} + \bar{W}_{\Phi:I} \times \bar{W}_{\Phi:O} + \bar{W}_{\Phi:O} \times \bar{W}_{\Phi:C}\big), \label{eq:loss:orthogonal}
\end{equation}
where: $\gamma \in \mathbb{R}_+$ governs the influence of $\mathcal{L}_O$ on training; and $\bar{W}_{\Phi:V}$ is a row average of the product of all the layers going from the representation, $\Phi$, to the factor $V$ \citep{kuang2017treatment}-- i.e. $\bar{W}_{\Phi:V} = \text{\texttt{rowavg}}(\theta_\Phi^1 \times \dots \times \theta_\Phi^l \times \theta_V^1 \times \dots \times \theta_V^m)$ where the subscript denotes the relevant network, and the superscript denotes the layer of this network, $\theta$ indicate the parameters of the specific network, also note that the weights are absolute when calculating $\bar{W}_{\Phi:V}$. In the LSTM we compute these products for every gate distinctly, and sum them together to match the input dimensions of the subsequent factor networks. As each $\bar{W}_{\Phi:V} \in \mathbb{R}^d$, we interpret the value of the $v$\textsuperscript{th} dimension in $\bar{W}_{\Phi:V}$ as the influence of the corresponding $v$\textsuperscript{th} dimension in $\Phi$ on the factor $V$. To gain some intuition in $\mathcal{L}_O$, imagine some dimensions in $\Phi$ to have high influence on two different factors. When we multiply these influence vectors with one another, the resulting loss will be high. Conversely, when alternate dimensions in two influence vectors are high and low (take zero as an extreme), the corresponding dimensions will be cancelled out, resulting in a low loss. As such, minimising $\mathcal{L}_O$ results in regions in $\Phi$ to have concentrated influence on less factors. We combine the above components in one loss function for joint optimisation,
\begin{equation}
    \mathcal{L} = \mathcal{L}_Y + \mathcal{L}_D + \mathcal{L}_C + \mathcal{L}_O + \mathfrak{R}, \label{eq:loss}
\end{equation}
where $\mathfrak{R}$ is a regularizer on the weights of our model, and $\mathcal{L}$ is a loss over all parameters, implying joint optimisation. In our experiments, we include ablation benchmarks where we cancel terms of eq.(\ref{eq:loss}). For a detailed overview of our optimisation procedure, we refer to our supplemental materials.

 \label{sec:experiments:synth}
\begin{figure}[t]
    \begin{subfigure}[c]{0.22\textwidth}
        \includegraphics[width=\textwidth]{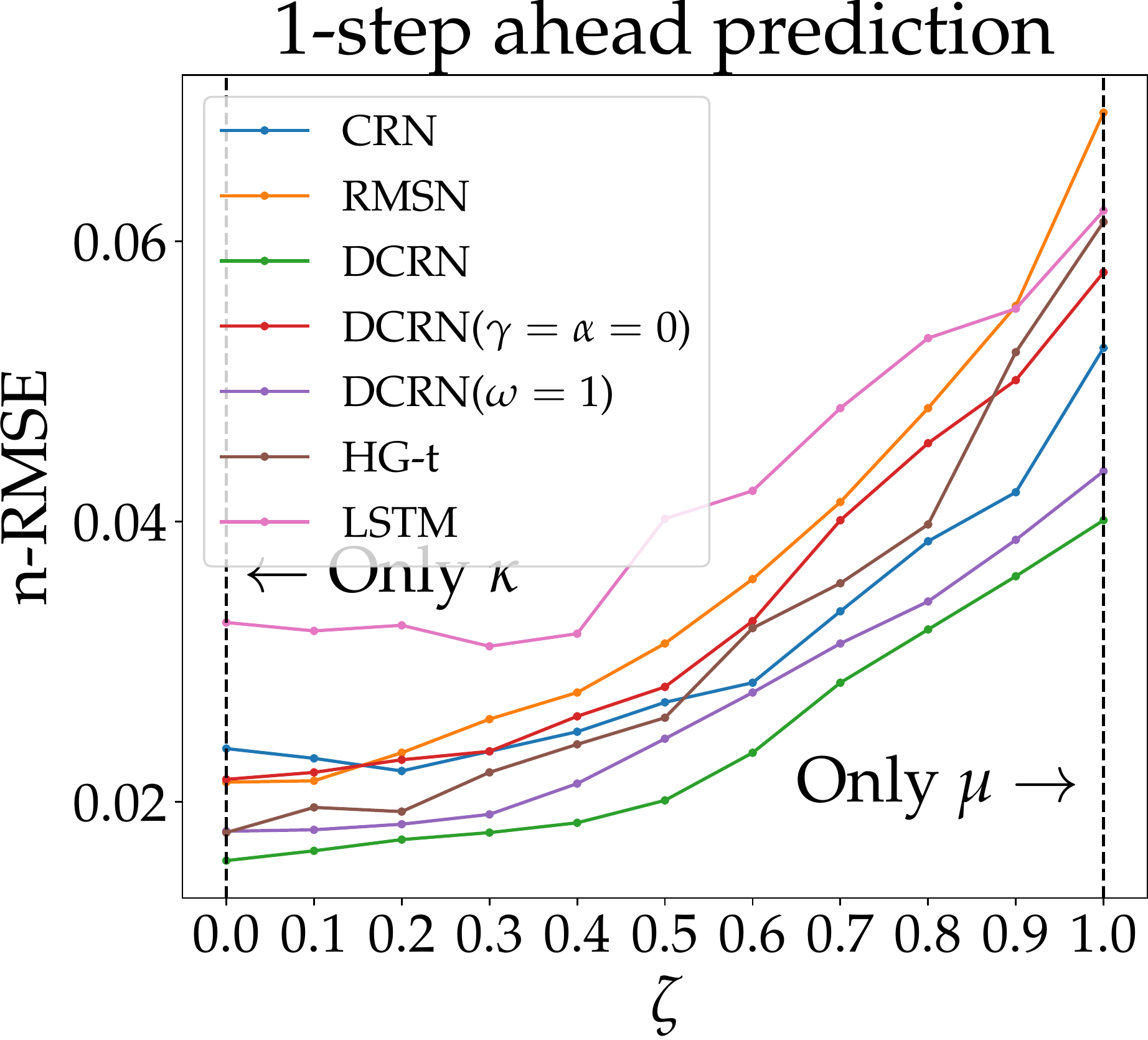}
    \end{subfigure}%
    ~
    \begin{subfigure}[c]{0.22\textwidth}
        \includegraphics[width=\textwidth]{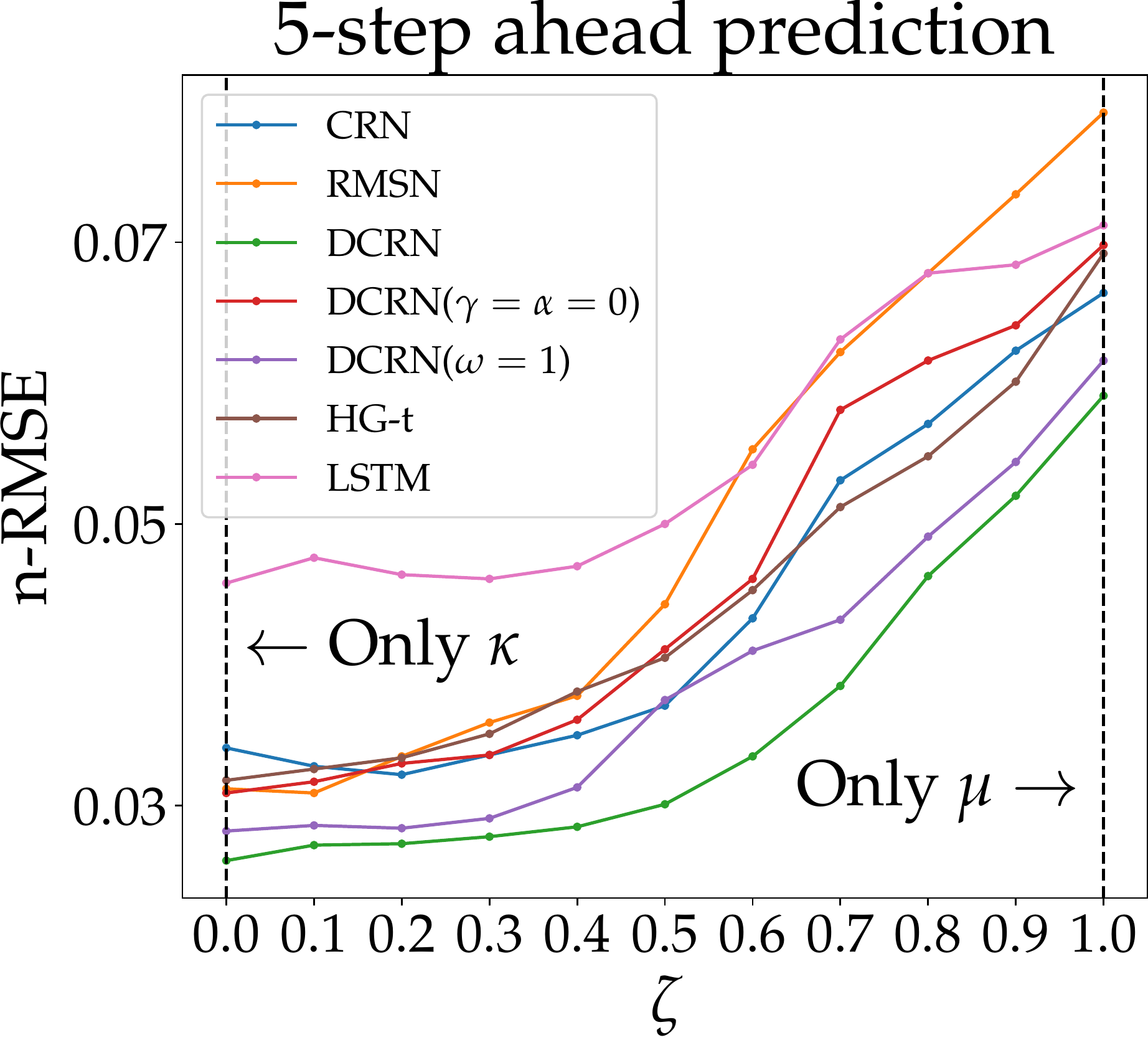}
    \end{subfigure}%
    \hfill
    \begin{subfigure}[c]{0.54\textwidth}
        \includegraphics[width=0.9\textwidth]{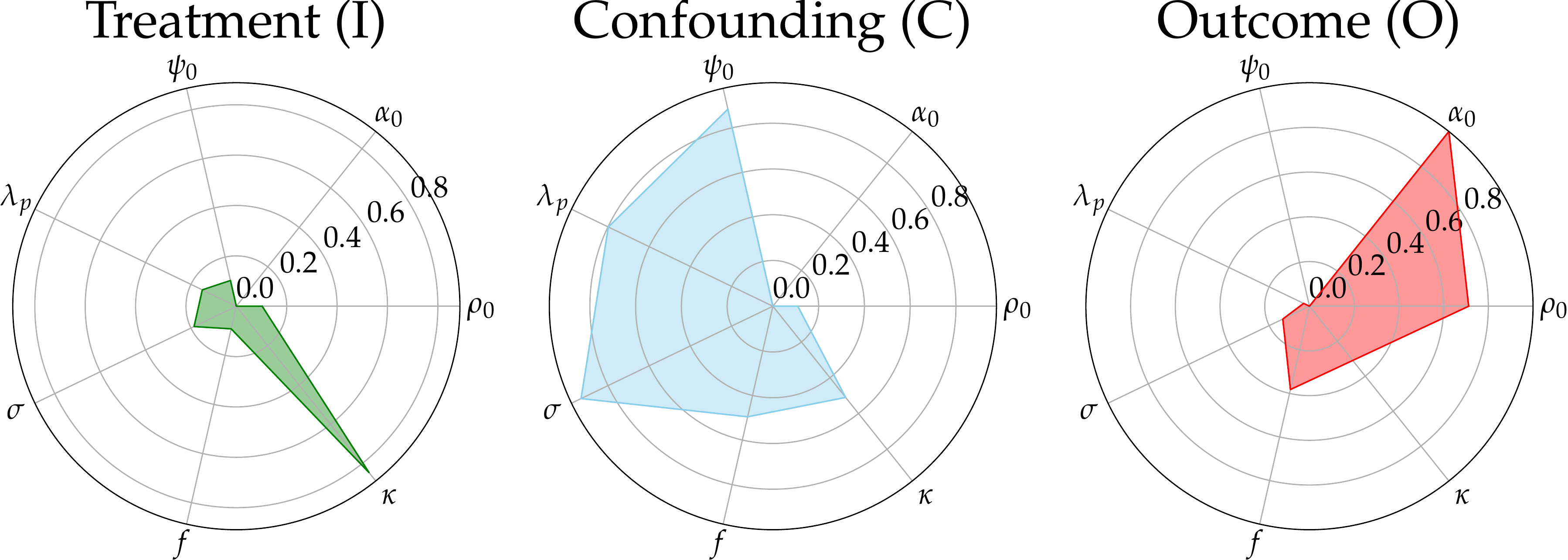}
    \end{subfigure}%
    \caption{{\bf Performance of DCRN on synthetic data.} {\it The leftmost two panels.} We report the 1-step ahead, and 5-step ahead counterfactual forecasting in the first and second panels, respectively. Performance on counterfactual $\tau$ step-ahead forecasting is expressed in the normalised RMSE (lower is better). {\it Rightmost three radar plots.} We have computed $\bar{W}_{\mathcal{X}:I}$, $\bar{W}_{\mathcal{X}:C}$, and $\bar{W}_{\mathcal{X}:O}$ with $\mathcal{X}$ the patient covariates in \citep{butner2020mathematical} and $\zeta=0.7$. From these results we learn that more influential variables in $\mathcal{X}$ (those that are further from the center), correctly correspond to those defined in our synthetic setup.}
    \label{fig:results:synth:tau-step}
    \rule{\linewidth}{.75pt} 
    \vspace{-7mm}
\end{figure}

\section{Experiments} \label{sec:experiments}
With DCRN our goal is to perform accurate forecasting, as well as to gain some understanding on how patient covariates influence their trajectories in terms of treatment selection and outcome. An important consideration for both objectives is the role of a ground truth data-generating process: for forecasting, we are interested in counterfactual performance, i.e. evaluating DCRN's ability to forecast patient outcomes for treatment strategies different than the ones present in the dataset; for factor analysis, we want to compare the factors DCRN has learned to the actual influence of the covariates on the factors, which is generally unknown in real datasets. Therefore, we test our model on synthetic data, where  the ground truth is known, as well as real data, where we can compare with only factual outcomes.

\textbf{Synthetic data.} We use \citet{butner2020mathematical}'s PKPD (pharmacokinetic-pharmacodynamic)  model of tumour growth under immune therapy, where tumour growth is modelled as a differential equation,
\begin{equation}
    \frac{d\rho}{dt} = (\alpha_0 - \mu + \mu \cdot \Lambda) \cdot \rho - \mu \cdot \Lambda \cdot \rho^2 \label{eq:master_equation},
\end{equation}
where: $\rho$ is the cancer cell mass; $\alpha_0$ is the tumour proliferation rate; $\mu$ is the tumour cell killing rate, computed as $\mu = \lambda_p \cdot \psi_0 \cdot \sigma$, with $\sigma$ the drug bound to the immune cells, $\psi_0$ the immune cell counts, and $\lambda_p$ the intrinsic kill rate of immune cell therapy; $\Lambda = \frac{\rho_0/\psi_0}{f}$ with $\rho_0$ the tumour cell count and $f$ the immune-cell fitness, represents a measure of the anti-tumour immune state. For one time-step: we update the cancer cell mass, $\rho$, by means of eq.(\ref{eq:master_equation}), and we select a treatment, discussed below.

\cite{butner2020mathematical}'s PKPD model describes a patient trajectory under constant treatment, however, we are interested in a model where the patient's trajectory is determined by a per time-step treatment decision. As such, from eq.(\ref{eq:master_equation}) we learn that higher tumour cell killing rate might decrease tumour size, and thus increase this kill rate when treated and decrease it when left untreated; specifically we alter $\psi_0$-- the immune cell count. The treatment decision then, is based on a Bernouilli random variable, with probability $p_a$, depending on the cancer cell mass, as well as an unrelated autoregressive process: $p_a(t) = \text{sigmoid}(\zeta\bar{\mu}(t) + (1-\zeta)(\kappa(t)))$ and $A_t \sim \text{Bernoulli}(p_a(t))$, with $\bar{\mu}(t)$ the tumour cell killing rate of the last $10$ time-steps--- i.e., when the treatment appears effective, more is given ---and $\kappa(t)$ a simple autoregressive process alongside eq.(\ref{eq:master_equation}), making treatment selection a time-dependent random variable. As $\kappa(t)$ is a complete distinct process from cancer progression, it acts as a treatment factor. With $\zeta \in [0, 1]$ we can completely remove the influence of this treatment factor, or make it dominant. More information regarding this model can be found in our supplemental material, as well as \citep{butner2020mathematical}.

\textbf{Real data.} We use the Medical Information Mart for Intensive Care (MIMIC-III) \cite{johnson2016mimic}. From MIMIC-III we use the temporal features to predict white blood cell count with antibiotics as the treatment. A similar setup was used in \citep{bica2020estimating}: we extract the patients with up to 30 timsteps (3487 patients in total); with 26 patient covariates measured over time.

\textbf{Benchmarks.} There are two benchmarks of relevance to our work: Counterfactual Recurrent Networks (CRNs) \citep{bica2020estimating}, and Recurrent Marginal Structure Networks (RMSNs) \citep{lim2018forecasting}. Furthermore, we also compare against three ablations: one where we set $\gamma = \alpha = 0$, thus removing $\mathcal{L}_O$ and $\mathcal{L}_D$ from the loss function (DCRN($\gamma=\alpha=0$)); one where we set $\omega=1$, to measure the influence of weighting the forecasts in our loss functions (DCRN($\omega=1$)); and one where we naivly extend \citet{hassanpour2020} to the temporal setting with three distinct recurrent networks (HG-t). All models were tuned with a Bayesian hyperparameter optimisation scheme, described in the supplemental material.

\subsection{Experiments on synthetic data}

\textbf{Forecasting.} With our synthetic data-generating process described above, we can create observational datasets with various degrees of time-dependent confounding bias. Once a model is trained using this biased dataset, we give a patient history equivalent to those present in the training dataset, but generate for one patient multiple counterfactual treatment trajectories for the last $\tau$ steps. In doing so, we can evaluate how well a model is able to make out-of-sample forecasts as these counterfactual trajectories contain treatment strategies different than the ones represented in the training data. In our supplemental materials, we have provided a more detailed overview on our test dataset.

In our simulation, $\zeta$ controls how much influence the treatment factor expresses on treatment selection compared to the confounding factor, $\mu$. As such, we have generated eleven datasets where we vary $\zeta$ as $0.0, 0.1, \dots, 1.0$, and report the performance of DCRN as well as our benchmarks. We have reported results on one step-ahead, and five step-ahead forecasting as a function of $\zeta$ in Figure~\ref{fig:results:synth:tau-step}. In our results, we report RMSN to have worst performance, as was also reported in \cite{bica2020estimating}. We also notice that CRN's performance deteriorates somewhat when the treatment factor expresses most influence on treatment selection. We find this to be a logical result as CRN's balancing mechanism functions as a regularizer over the entire input space, contrasting DCRN as well as RMSN, which employ weighting. We also report good performance on the naive extension of \citet{hassanpour2020}, as the only other method to disentangle factors, this is to be expected. Furthermore, we report better performance across all methods when the treatment factor is most dominant, the reason for this is likely due to the lower dimensionality of $\kappa$ compared to $\mu$, i.e. resulting in less profound selection bias. Next to prediciton accuracy, we also test DCRN in composing effective treatment plans. In Table~\ref{tab:results:mimic}, we show on synthetic data that DCRN was able to select the optimal treatment plan 87\% and 80\% of the time, for 3-step and 5-step ahead, respectively--- an important quality in practice.

\textbf{Factor analysis.} Next to accurate forecasts, we are also interested in whether the disentangled factors hold meaningful information with respect to the input space. As such, we have visualised the influence vectors, $\bar{W}_{\mathcal{X}:I}$, $\bar{W}_{\mathcal{X}:C}$, and $\bar{W}_{\mathcal{X}:O}$, representing the influence the input space has on the treatment, confounding, and outcome factor respectively in Figure~\ref{fig:results:synth:tau-step}~(right). In our results we report that the terms from \cite{butner2020mathematical}'s PKPD model are correctly categorised by DCRN. Specifically, we recognize $\kappa$ as the only component in the treatment factor; $\lambda_p$, $\sigma$, and $\psi_0$ as the components making up $\mu$; and $\alpha_0$ with $\rho_0$ as the outcome factors. While $\mathcal{L}_O$ enforces hard disentanglement, there is still room for error, illustrated by $\kappa$ and $f$ having \textit{some} influence on the confounding factor. The reason for this is that $\mathcal{L}_O$ should be regarded as a regulariser, which is usually toned down by setting $\gamma$ to some value less than $1$. Naturally, this factor analysis can also be provided for the real-world MIMIC-III data, for this we refer to \cref{sec:app:mimic}, for a full clinical analysis of our found factors.

\subsection{Experiments on real data}
\textbf{Forecasting.} Similar to our synthetic setup above, we train DCRN and the benchmarks with a biased training dataset. For testing, we iterate over every patient sequence and test every model's forecast for the next $\tau$ timesteps on factual outcomes. While this experiment offers some assurance on whether our method is able to handle real data (MIMIC, specifically \citep{johnson2016mimic}), it does not test for counterfactual forecasting as the test dataset is equally biased as the training dataset, i.e. there are no out-of-sample data in the testing dataset. The reason is simply due to the absence of a ground truth to construct counterfactual trajectories. From Table~\ref{tab:results:mimic}, we learn that DCRN performs well on real data.

\begin{table}[t]
    \centering
    \caption{{\bf Results on forecasting, and treatment selection.} First we report the n-RMSE and standard deviation on $\tau$-step ahead forecasting of factual outcomes for white blood cell count (outcome) after antibiotics (treatment) in MIMIC-III, given 10 different folds of the data. Next, we also report good performance on the accuracy of optimal treatment selection using synthetic data.}
    \begin{tabular}{r c c | c c c}
        \toprule
        & \multicolumn{2}{c}{\bf Optimal treatments (Synth.)} & \multicolumn{3}{c}{\bf Forecasting performance (Real)} \\
        & 3-step & 5-step & 1-step & 3-step & 5-step  \\
        \midrule
        CRN & 0.84 \scriptsize{(0.01)} & 0.78 \scriptsize{(0.00)} & 3.1531\scriptsize{(0.3033)} & 3.3656 \scriptsize{(0.0825)} & 4.2795 \scriptsize{(0.0745)}  \\
        RMSN & 0.81 \scriptsize{(0.02)}  & 0.72 \scriptsize{(0.02)} & 3.4034 \scriptsize{(0.2101)} & 3.6708 \scriptsize{(0.2310)} & 5.1986 \scriptsize{(0.1897)}  \\
        \rowcolor{lightcyan!60} DCRN &    {\bf 0.87} \scriptsize{(0.04)} & {\bf 0.80} \scriptsize{(0.03)} & {\bf 2.2454} \scriptsize{(0.5603)} & {\bf 2.8795} \scriptsize{(0.4829)} & {\bf 3.7070} \scriptsize{(0.4633)} \\
        \bottomrule
    \end{tabular}
    \label{tab:results:mimic}
    \vspace{-5mm}
\end{table}

\vspace{-2mm}
\section{Conclusion} \label{sec:conclusion}
\vspace{-2mm}
We advance  temporal individual treatment effect (ITE) estimation. DCRN explicitly disentangles patient covariates into a treatment factor, an outcome factor, and a confounding factor. Through the disentanglement of DCRN, we can gain some understanding of how different patient covariates influence outcomes and treatment selection in the data. Such interpretation could prove valuable for many clinicians, trying to understand various dynamics of different diseases and their accompanying treatment plans. One limitation of our method is that if our assumptions are violated, then DCRN will wrongly estimate any causal effect. Moreover, as with any other methods for forecasting treatment outcomes over time, DCRN could potentially have negative societal impact if someone uses it to select treatment plans resulting in worse patient outcomes.


\bibliographystyle{unsrtnat}
\bibliography{example_paper}


\newpage
\appendix
\section{DCRN Optimisation procedure} \label{app:opt}
Recall from our main text, there are two phases to train DCRN: (i) train the encoder; (ii) use the encoder to encode the patient histories, used to train the decoder. As such, we present our training procedure in two main blocks, illustrated in Algorithm~\ref{alg:full}. With Algorithms~\ref{alg:encoder}~and~\ref{alg:decoder}, we provide further details on training the encoder and decoder, respectively. Note that training DCRN requires backpropagating over multiple networks jointly. For DCRN (in the PKPD setup) the average runtime is \texttildelow 8min (Nvidia GeForce RTX 2080 Ti) with \texttildelow 700k parameters.

\begin{algorithm}[h]
    \caption{DCRN complete training procedure.}\label{alg:full}
    
    \SetAlgoLined
    \SetKwInOut{Input}{Input}
    \SetKwFunction{EncoderGradientStep}{EncoderGradientStep}
    \SetKwFunction{DecoderGradientStep}{DecoderGradientStep}
    \SetKwFunction{Validate}{Validate}
    \SetKwFunction{EarlyStop}{EarlyStop}
    
    \Input{Training data,\\\quad $\mathcal{T} = \{\mathbf{X}^{(i)}_t, Y^{(i)}_t, A^{(i)}_t\}_{i=1,\dots,N; t=1,\dots,T^{(i)}}$\\Validation data,\\\quad $\mathcal{V} = \{\mathbf{X}^{(j)}_u, Y^{(j)}_u, A^{(j)}_u\}_{j=1,\dots,M; u=1,\dots,U^{(j)}}$\\ Untrained encoder, \texttt{enc}\\ Untrained decoder, \texttt{dec}}
    \tcc{BLOCK 1: train encoder.}
    \For{epochs}{
        \For{$\mathcal{B} \sim \mathcal{T}$}{
            \texttt{enc} $\leftarrow$ \EncoderGradientStep{$\mathcal{B}$}\;
        }
        $l_v \leftarrow$ \Validate{\texttt{enc}, $\mathcal{V}$}\;
        \EarlyStop{\texttt{enc}, $l_v$}\;
    }
    \tcc{Fix parameters of \texttt{enc}.}
    \tcc{BLOCK 2: train decoder.}
    \For{epochs}{
        \For{$\mathcal{B} \sim \mathcal{T}$}{
            \texttt{dec} $\leftarrow$ \DecoderGradientStep{$\mathcal{B}$}\;
        }
        $l_v \leftarrow$ \Validate{\texttt{dec}, $\mathcal{V}$}\;
        \EarlyStop{\texttt{dec}, $l_v$}\;
    }

\end{algorithm}

\begin{algorithm*}[t]
    \SetAlgoLined
    \SetKwInOut{Input}{Input}
    \SetKwFunction{RNNA}{$\text{RNN}_A^E$}
    \SetKwFunction{RNNP}{$\text{RNN}_\Phi^E$}
    \SetKwFunction{RNNY}{$\text{RNN}_Y^E$}
    \SetKwFunction{I}{$\text{I}^E$}
    \SetKwFunction{C}{$\text{C}^E$}
    \SetKwFunction{O}{$\text{O}^E$}
    \SetKwFunction{AA}{$\text{A}^{E, *}$}
    \SetKwFunction{A}{$\text{A}^E$}
    \SetKwFunction{Y}{$\text{Y}^E$}
    \SetKwFunction{GradientStep}{GradientStep}
    
    \Input{Minibatch, $\mathcal{B} = \{\mathbf{X}^{(i)}_t, Y^{(i)}_t, A^{(i)}_t\}_{i=1,\dots,B; t=1,\dots,T^{(i)}}$}

    \For{$t$ in $1, 2, \dots, T$}{
        \tcc{sample data from minibatch.}
        $\mathbf{X}_t, A_t, Y_t \sim \mathcal{B}$\;
        
        \tcc{encode data.}
        $\Phi_t \leftarrow$ \RNNP{$\mathbf{X}_t$, $h_X$}\;
        $\mathbf{H}_{A_t} \leftarrow$ \RNNA{$A_t$, $h_A$}\;
        $\mathbf{H}_{Y_{t+1}} \leftarrow$ \RNNY{$Y_{t+1}$, $h_Y$}\;
        
        \tcc{create factors.}
        $I_t \leftarrow$ \I($\Phi_t$)\;
        $C_t \leftarrow$ \C($\Phi_t$)\;
        $O_t \leftarrow$ \O($\Phi_t$)\;
        
        \tcc{make predictions.}
        $\hat{A}_t \leftarrow$ \A{$I_t$, $C_t$}\;
        $\hat{A}^*_t \leftarrow$ \AA{$I_t$}\;
        $\hat{Y}_{t+1} \leftarrow$ \Y{$C_t$, $O_t$, $\mathbf{H}_{Y_{t+1}}$, $\mathbf{H}_{A_t}$, $A_t$}\;
        
        \tcc{calculate loss.}
        $l_e \leftarrow \mathcal{L}(\hat{A}_t, \hat{A}^*_t, \hat{Y}_{t+1}, \text{\texttt{RNN}}_\Phi^E, \text{\texttt{RNN}}_A^E, \text{\texttt{RNN}}_Y^E, \text{\texttt{Y}}^E, \text{\texttt{A}}^E, \text{\texttt{A}}^{E, *}, \text{\texttt{I}}^E, \text{\texttt{C}}^E, \text{\texttt{O}}^E)$\;
        
        \tcc{take gradient step.}
        \texttt{enc}$\leftarrow$ \GradientStep{$l_e$, \texttt{enc}}\;
    }
    
    \caption{\texttt{EncoderGradientStep} in detail.}\label{alg:encoder}
\end{algorithm*}

\begin{algorithm*}[t]
    \SetAlgoLined
    \SetKwInOut{Input}{Input}
    \SetKwFunction{RNNA}{$\text{RNN}_A^D$}
    \SetKwFunction{RNNP}{$\text{RNN}_\Phi^D$}
    \SetKwFunction{RNNY}{$\text{RNN}_Y^D$}
    \SetKwFunction{I}{$\text{I}^D$}
    \SetKwFunction{C}{$\text{C}^D$}
    \SetKwFunction{O}{$\text{O}^D$}
    \SetKwFunction{AA}{$\text{A}^{D, *}$}
    \SetKwFunction{A}{$\text{A}^D$}
    \SetKwFunction{Y}{$\text{Y}^D$}
    
    \Input{Minibatch, $\mathcal{B} = \{\mathbf{X}^{(i)}_t, Y^{(i)}_t, A^{(i)}_t\}_{i=1,\dots,B; t=1,\dots,T^{(i)}}$}

    \For{$t$ in $1, 2, \dots, T-\tau$}{
        \tcc{sample data from minibatch.}
        $\mathbf{X}_t, A_t, Y_t \sim \mathcal{B}$\;
        
        \tcc{encode data using trained encoder.}
        \tcc{get one step-ahead prediction using trained encoder.}
        $\Phi_t, \mathbf{H}_{A_t}, \mathbf{H}_{Y_{t+2}}, \hat{Y}_{t+2} \leftarrow \text{\texttt{enc}}(\mathbf{X}_{1:t}, A_{1:t-1}, Y_{1:t+1})$\;
        
        \tcc{set autoregressive variables.}
        
        \For{$u$ in $\tau$}{
            \tcc{build $\tau$ step-ahead predictions.}
            
            \tcc{encode data.}
            $\Phi_{u + T} \leftarrow$ \RNNP{$\Phi_r$} \;
            $\mathbf{H}_{A_r} \leftarrow$ \RNNA{$A_{u + T}$} \;
            $\mathbf{H}_{Y_r} \leftarrow$ \RNNA{$\hat{Y}_{r}$} \;
            
            \tcc{create factors.}
            $I_{u + T} \leftarrow$ \I{$\Phi_r$} \;
            $C_{u + T} \leftarrow$ \C{$\Phi_r$} \;
            $O_{u + T} \leftarrow$ \O{$\Phi_r$} \;
            
            \tcc{make predictions.}
            $\hat{A}_{u+T} \leftarrow$ \A{$I_{u+T}$, $C_{u+T}$} \;
            $\hat{A}_{u+T}^* \leftarrow$ \AA{$I_{u+T}$} \;
            $\hat{Y_{u + T + 1}} \leftarrow $ \Y{$C_{u+T}$, $O_{u+T}$, $\mathbf{H}_{Y_r}$, $\mathbf{H}_{A_r}$, $A_{u + T}$} \;

        }
        \tcc{calculate losses and take gradient step.}
        $l_d \leftarrow \mathcal{L}(\hat{A}_{T:T+\tau-1}, \hat{A}^*_{T:T+\tau-1}, \hat{Y}_{T:T+\tau}, \text{\texttt{RNN}}_\Phi^E, \text{\texttt{RNN}}_A^E, \text{\texttt{RNN}}_Y^E, \text{\texttt{Y}}^E, \text{\texttt{A}}^E, \text{\texttt{A}}^{E, *}, \text{\texttt{I}}^E, \text{\texttt{C}}^E, \text{\texttt{O}}^E)$\;
        Take gradient step w.r.t., $l_d$\;
    }

    \caption{\texttt{DecoderGradientStep} in detail.}\label{alg:decoder}
\end{algorithm*}

\section{Hyperparameters and benchmarks} \label{app:hyper}
For RMSN and CRN we relied on Clairvoyance \citep{jarrettclairvoyance}. While Clairvoyance implements a hyperparameter optimisation scheme for RMSN, it does not for CRN. As such, we have performed our own hyperparameter search for CRN, alongside our hyperparameter search for DCRN. For CRN we adopted the same ranges as reported in \cite{bica2020estimating} when applicable, also described in leftmost side of Table~\ref{tab:hp:CRN}. For DCRN, we report the hyperparameter ranges in the rightmost side of Table~\ref{tab:hp:CRN}. Both tables contain the eventual optimal parameters. Note that, we have searched over a generated dataset with $\zeta=0.5$ and kept these parameters throughout the experiment for computational considerations. Note that the performance of the hyperparameters is evaluated on factual data, as we generally have no access to counterfactual data.

\begin{table*}[t]
    \centering
    \caption{Hyperparameter search ranges for CRN and DCRN, values in brackets indicate the chosen parameter value, dashes indicate the total continuous range as we use Bayesian optimisation for the search method. For each component, we ran 20 iterations. Values are rounded to sensible values, e.g. a learning rate of 0.00123286 is reported as 0.001 for brevity.}
    \begin{tabular}{r c c c c}
         \toprule
          &  \multicolumn{2}{c}{CRN} & \multicolumn{2}{c}{DCRN} \\
          \cmidrule(lr){2-3} \cmidrule(lr){4-5}
          Hyperparameter & Encoder & Decoder & Encoder & Decoder \\
         \midrule
         Learning rate & \footnotesize{0.01 - 0.0001 (0.001)} & \footnotesize{0.01 - 0.0001 (0.001)} & \footnotesize{0.01 - 0.0001 (0.001)} & \footnotesize{0.01 - 0.0001 (0.001)}\\
         Minibatch size & \footnotesize{64 - 256 (64)} & \footnotesize{256 - 512 (128)} & \footnotesize{16 - 256 (128)} & \footnotesize{64 - 512 (256)}\\
         RNN hidden units & \footnotesize{8 - 128 (16)} & \footnotesize{8 - 128 (16)} & \footnotesize{8 - 128 (16)} & \footnotesize{8 - 128 (16)} \\
         Representation size & \footnotesize{16- 256 (16)} & \footnotesize{16- 256 (32)} & \multicolumn{2}{c}{\footnotesize{8 - 256 (16)}} \\
         FC hidden units & \footnotesize{4 - 32 (24)}  & \footnotesize{4 - 32 (24)} & \footnotesize{4 - 32 (16)} & \footnotesize{4 - 32 (16)} \\
         RNN dropout & \footnotesize{0 - 0.4 (0.1)} & \footnotesize{0 - 0.4 (0.1)} & \footnotesize{0 - 0.4 (0.1)} & \footnotesize{0 - 0.4 (0.1)} \\
         $\alpha$ & n.a. & n.a & \footnotesize{0 - 1 (0.4)} & \footnotesize{0 - 1 (0.4)} \\
         $\beta$ & n.a. & n.a. & \footnotesize{0 - 1 (1)} & \footnotesize{0 - 1 (1)}\\
         $\gamma$ & n.a. & n.a. & \footnotesize{0 - 1 (0.3)} & \footnotesize{0 - 1 (0.3)} \\
         \bottomrule
    \end{tabular}
    
    \label{tab:hp:CRN}
\end{table*}

\section{Factor analysis on MIMIC-III} \label{sec:app:mimic}
\begin{figure}
    \centering
    \includegraphics[width=.6\textwidth]{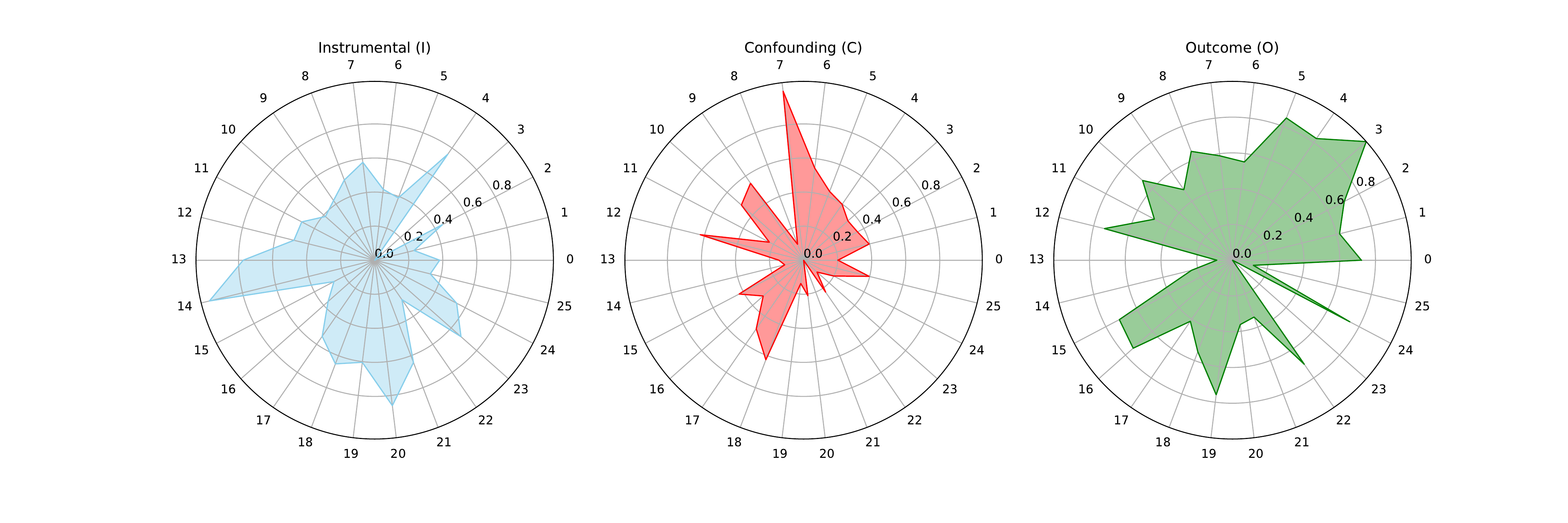}
    \caption{{\bf Factor analysis of MIMIC-III data.} The same analysis as in \Cref{fig:results:synth:tau-step}, executed on the representations learnt for MIMIC-III. For a clinical interpretations we refer to \Cref{sec:app:mimic}.}
    \label{fig:res:factor_MIMIC}
    \rule{\linewidth}{.75pt}
\end{figure}

In DCRN, the longitudinal trajectory is decomposed into three latent factors that vary over time and influence either treatment (I, Antibiotic therapy for the MIMIC dataset), outcome (O, white cell count in MIMIC-III) or both (as a confounding factor, C). By decomposing inputs that influence only treatment (I) or outcome (O) DCRN creates the potential for us to gain insight into the mechanism by which each process is occurring. This is a specific advantage of the latent factorisation approach taken by DCRN. 

Analysing the weighted contribution of each covariate to each factor in the MIMIC-III dataset as in Figure~\ref{fig:res:factor_MIMIC} (analogous to Figure~\ref{fig:results:synth:tau-step}) creates the potential to identify previously unsuspected clinical covariates that are important influences on either the clinical decision to give antibiotic therapy (factor $I$) or the outcome resulting from that action (factor $O$). For the treatment factor influencing antibiotic therapy our analysis of the MIMIC-III dataset identifies strong contribution from ‘Mean blood pressure’, ‘Systolic blood pressure’ (Sys. BP)), ‘Respiratory rate’ (Resp. rate), ‘INR’, and ‘Creatinine’. These covariate weightings make coherent sense with each known to be not just associated with clinical sepsis but forming part of the diagnostic criteria \cite{rhodes2017surviving}. In sepsis, hypotension resulting from blood vessel dilatation results in hypoperfusion of many tissues with reduced kidney function \cite{alobaidi2015sepsis} (manifest as an elevated creatinine level) and a sepsis-associated coagulopathy \cite{simmons2015coagulopathy} (manifest as an increase in INR or PTT value) as characteristic features. 

The strongest covariate weights contributing to the instrumental factor (I) from DCRN therefore comprise a coherent set of established clinical traits known to be associated with sepsis. This reinforces the potential utility of the method although arguably does not advance our understanding of sepsis. 
However, although much attention has been devoted to early identification and diagnosis of sepsis \cite{fleuren2020machine} as this can improve mortality, predicting outcome (in this case response to antibiotic therapy) is more challenging.  Both predicting and understanding response to antibiotic therapy would be useful advances but are complicated by several confounding factors in clinical practice \cite{vincent2016clinical}. Underlying inflammatory conditions can mimic sepsis \cite{gando2020sirs}, giving a systemic inflammatory response syndrome (SIRS) that is neither driven by infection nor responsive to antibiotic therapy. Sepsis is a complex multisystem disorder that can be caused by myriad pathogens -  including those with intrinsic or acquired antibiotic resistance – and with infection in a disparate range of anatomical sites. It is clear that not all ICU cases meeting the diagnostic criteria for SIRS will have sepsis and not all those who have sepsis will respond to antibiotic therapy. Decomposing the contribution of weighted covariates contributing to the outcome factor in DCRN could offer insight into this process. The majority of strong covariate weights contributing to the outcome factor from DCRN applied to the MIMIC-III dataset are associated with severity of sepsis, including blood pressure, temperature and measures of kidney function (bun and creatinine), known to be associated with sepsis outcome \cite{peerapornratana2019acute, charles2014predicting}. The strongest weighted covariate however – chloride level - was more unexpected. This does not form part of the criteria for sepsis or the definition of severe sepsis although its level does associate with the degree of sepsis-associated acute kidney injury \cite{suetrong2016hyperchloremia}. It has however, been associated with mortality in sepsis \cite{oh2017increased} and its identification here reinforces that association and suggests that it may, along with more accepted measures of sepsis severity, be linked to treatment response.

\begin{table}[h]
    \centering
    \caption{{\bf Temporal variables in MIMIC-III.} In Figure~\ref{fig:res:factor_MIMIC} each variable is assigned a number from $1,2,\dots,26$, which correspond to the variables below.}
    \label{tab:MIMIC:variables}
    \begin{tabular}{rl | rl}
    \toprule
    {\bf Code} & {\bf Variable} & {\bf Code} & {\bf Variable}\\
    \midrule
         0 & Time & 13 & INR\\
         1 & Bicarbonate& 14 & Mean blood press.\\
         2 & BUN& 15 & Oxygen Therapy\\
         3 & Chloride& 16 & Platelet\\
         4 & Creatinine& 17 & Potassium\\
         5 & DiasBP& 18 & PT\\
         6 & Extubated& 19 & PTT\\
         7 & FIO2& 20 & Resp. rate\\
         8 & Glucose chart& 21 & Sodium\\
         9 & Glucose lab& 22 & SPO2\\
         10 & Heartrate high& 23 & Sys. BP\\
         11 & Hematocrit& 24 & Temperature\\
         12 & Hemoglobin& 25 & Ventilator\\
    \bottomrule
    \end{tabular}
\end{table}

\section{Impulse response modelling}
Contrasting CRN, DCRN learns not only how the patient outcome evolves over time, but also treatment assignment. This allows us to consider the problem of forecasting patient responses $\tau$ time steps ahead both by specifying a $\tau$-step treatment sequence \textit{and} by specifying only an initial intervention, considering the forecasting problem as a generalised impulse response problem \cite{koop1996impulse}. 

Specifically, because our model also learns an autoregressive model of treatment assignment, we could also consider the response to a single treatment $A_t=\tilde{a}_t,$, i.e.  $\{\mathbbm{E}[Y(\tilde{a}_t)_{t+\tau}|H_t]\}_{\tau \geq 1} = \{\mathbbm{E}[Y_{t+\tau}|A_t=\tilde{a}_t, H_t]\}_{\tau \geq 1} $, as a generalised impulse-response analysis problem \citep{koop1996impulse}. That is, we could consider the setting where we provide the dynamical system with an exogenous (treatment) impulse resulting in treatment assignment $a$ at time $t$, and are interested in the response -- the development of the patient state (given its history) should no further shocks (to outcomes $Y_{t+\tau}$, covariates $X_{t+\tau}$ and treatment process $A_{t+\tau}$ for $\tau\geq 1$) hit the system, where the system ``completes" the treatment-plan given all present treatment-selection bias. We defer analysis of this setting to future work.

\section{Details on data simulation}
We describe the data-generating process as per \cite{butner2020mathematical}.
\subsection{Synthetic data-generation}
Every patient has eight dimensions, let a patient be sampled as: $\alpha_0$, the tumour proliferation rate, which we sample from $\alpha_0 \sim \mathcal{U}(0, 0.1)$; $\lambda_p$, the intrinsic kill rate of immune cell therapy, sampled from $\lambda_p \sim \mathcal{U}(0.1, 0.2)$; $\sigma$, the drug bound to the immune cells, sampled as $\sigma \sim \mathcal{U}(0, 0.1)$; $\psi_0$ as the immune cell counts, sampled from $\psi_0 \sim \mathcal{U}(0.1, 0.3)$; $\rho_0$ as the tumour cell count, sampled from $\rho_0 \sim \mathcal{U}(0, 0.1)$; and $f$ the immune-cell fitness, sampled from $f \sim \mathcal{U}(1, 5)$. For these distributions we have copied the first and second moments of the corresponding variables as reported in \citep{butner2020mathematical}. The autoregressive process $\kappa(t)$ is defined as $\kappa(t) = \sum_{i=0}^p \theta_i \kappa(t-i) + \epsilon_\kappa$, with $\theta_i \sim \mathcal{N}(0, 0.01) \forall i$ in $p$. In our simulation we let $p=10$, to match the $10$-step average of $\bar{\rho}(t)$.

As such, for every patient we first sample the variables described above:

\begin{align*}
    \mathbf{X}(t=0) &= (\alpha_0, \lambda_p, \sigma, \psi_0(t=0), \rho_0, f, \kappa(t=0), \rho(t=0) )^\top,\\
    \kappa(t=0) &\sim \mathcal{N}(0, 0.01),\\
    \psi_0(t=0) &\sim \mathcal{U}(0.1, 0.3),\\
    \rho(t=0) &\sim \mathcal{U}(0.03, 1),
\end{align*}
and let the covariates change as,
\begin{align*}
    A_t &\sim \text{Bern}(p_a(t)),\\
    \bar{\mu}(t) &= \frac{1}{N}\sum_{i=0}^N\mu(t-i),\\
    \mu(t) &= \lambda_p \cdot \psi_0(t) \cdot \sigma,\\
    \psi_0(t) &= \begin{cases}
    \psi_0(t-1) \cdot 1.01 & \text{if } A_t = 1\\
    \psi_0(t-1) \cdot 0.99 & \text{if } A_t = 0
    \end{cases}
\end{align*}
with $\epsilon_\rho \sim \mathcal{N}(0, 0.01)$.

For every timestep, we use the time-dependent  $\bar{\mu}(t)$ and $\kappa(t)$ to update $\rho(t)$ as,
\begin{equation*}
    \rho(t+1) = \rho(t) + \big( (\alpha_0 - \mu(t) + \mu(t) \cdot \Lambda)  \cdot \rho(t) - \mu(t) \cdot \Lambda \cdot \rho(t)^2 \big) + \epsilon_\rho 
\end{equation*}

With above simulation, we have variables that: only affect treatment selection ($\kappa(t)$); only affect outcome ($\Lambda$, $\alpha_0$, $\rho(t)$); and those that affect both ($\mu(t)$), corresponding with the treatment, outcome, and confounding factor, respectively. Furthermore, some variables have: a temporal influence ($\rho(t)$, $\kappa(t)$, $\psi_0(t)$); and some don't ($\alpha_0$, $\lambda_p$, $\sigma$, $\rho_0$, $f$)

\subsection{Generating test-data}
\begin{wrapfigure}{r}{0.5\textwidth}
    \centering
    \begin{tikzpicture}[
            roundnode/.style={circle, draw=tablenode!60, fill=tablenode!10, very thick,  minimum size=4mm, inner sep=0},
            textnode/.style={rectangle, inner sep=0, minimum width=5mm}
        ]
            \node (DCRN-tag) at (3, 1) {DCRN:};
            \node[roundnode] (data) {$\scriptstyle\mathcal{D}$};
            \node[roundnode] (hist) at (1, 0) {$\scriptstyle \mathbf{H}$};
            \node[roundnode] (phi) at (2, 0) {$\scriptstyle \Phi$};
            
            \node[textnode] (inst) at (4, .3) {$\scriptstyle I$};
            \node[textnode] (conf) at (4, 0) {$\scriptstyle C$};
            \node[textnode] (outc) at (4, -.3) {$\scriptstyle O$};
            
            \node[roundnode] (prop) at (6, .3) {$\scriptstyle A$};
            \node[roundnode] (outcome) at (6, -.3) {$\scriptstyle Y$};
            
            \draw[-latex] (data.east) -- (hist.west);
            \draw[-latex] (hist.east) -- (phi.west);
            
            \draw[-latex, rounded corners] (phi.east) -- (2.5, 0) -- (3, .3) -- (inst.west);
            \draw[-latex, rounded corners] (phi.east) -- (conf.west);
            \draw[-latex, rounded corners] (phi.east) -- (2.5, 0) -- (3, -.3) -- (outc.west);
            
            \draw[-latex] (inst.east) -- (prop);
            \draw[-latex, rounded corners] (conf.east) -- (4.5, 0) -- (5, .3) -- (prop.west);
            \draw[-latex, rounded corners] (conf.east) -- (4.5, 0) -- (5, -.3) -- (outcome.west);
            \draw[-latex] (outc.east) -- (outcome);
            
            \draw[latex-] (hist.north west) to [out=135, in=90, looseness=1.3] (.5, 0) to [out=270, in=225, looseness=1.3] (hist.south west);
            
            \node (HGt-tag) at (3, -1.5) {HG-t \cite{hassanpour2020}:};
            \node[roundnode] (data_hg) at (0, -2.5) {$\scriptstyle\mathcal{D}$};
            
            \node[roundnode] (inst_hg) at (2, -2.2) {$\scriptstyle I$};
            \node[roundnode] (conf_hg) at (3, -2.5) {$\scriptstyle C$};
            \node[roundnode] (outc_hg) at (4, -2.8) {$\scriptstyle O$};
            
            \node[roundnode] (prop_hg) at (6, -2.2) {$\scriptstyle A$};
            \node[roundnode] (outcome_hg) at (6, -2.8) {$\scriptstyle Y$};
            
            \draw[-latex] (data_hg.east) -- (conf_hg.west);
            \draw[-latex, rounded corners] (data_hg.east) -- (.5, -2.5) -- (1, -2.2) -- (inst_hg.west);
            \draw[-latex, rounded corners] (data_hg.east) -- (.5, -2.5) -- (1, -2.8) -- (outc_hg.west);
            
            \draw[-latex] (inst_hg.east) -- (prop_hg.west);
            \draw[-latex] (outc_hg.east) -- (outcome_hg.west);
            \draw[-latex, rounded corners] (conf_hg) -- (4.5, -2.5) -- (5, -2.8) -- (outcome_hg.west);
            \draw[-latex, rounded corners] (conf_hg) -- (4.5, -2.5) -- (5, -2.2) -- (prop_hg.west);
            
            \draw[latex-] (inst_hg.north west) to [out=135, in=90, looseness=1.3] (1.5, -2.2) to [out=270, in=225, looseness=1.3] (inst_hg.south west);
            \draw[latex-] (outc_hg.north west) to [out=135, in=90, looseness=1.3] (3.5, -2.8) to [out=270, in=225, looseness=1.3] (outc_hg.south west);
            \draw[latex-] (conf_hg.north west) to [out=135, in=90, looseness=1.3] (2.5, -2.5) to [out=270, in=225, looseness=1.3] (conf_hg.south west);
            
    \end{tikzpicture}

    \caption{{\bf Architecture comparison between DCRN and HG-t.} DCRN employs recurrent structures for the histories ($\mathbf{H}$), while a naive extension of \citet{hassanpour2020} (HG-t), employs a recurrent structure for each factor separately. Having recurrent structures on the ``factor-level", does not allow for the factors to influence each other over time.}
    \label{fig:HGt}
    \rule{\linewidth}{.75pt} 
\end{wrapfigure}
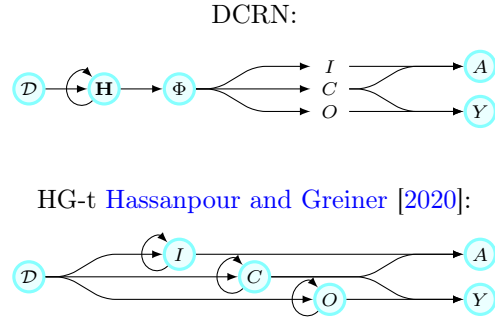
There are a few key points a good synthetic test for temporal ITE estimation should take into account: (i) the training data should be biased, as would be the case in a real environment; (ii) the test set should contain counterfactual trajectories, that is, the patient history should be equally biased as in the trainig set, but the future treatment trajectories should be unbiased. As such, for evaluating $\tau$-step ahead prediction, we sample for one patient a $T-\tau$ step trajectory, $\mathbf{H}_{\mathbf{X}_{T-\tau}}$, we copy this trajectory and complete it with one treatment in the following treatment plan, yielding $\tau$ counterfactual trajectories for every patient. We let $T$ vary from $1,2, \dots$, and let $\tau$ be as reported in the tables, indicating the amount of steps ahead the models predict. As an example, let $\max(T) = 20$ and $\tau = 5$ with $50$ patients, then the amount of trajectories in the testset is $20 \cdot 50 \cdot\tau = 5000$. 

Testing with this data involves feeding the first $T-\tau$ timesteps to the encoder model, and then let the decoder model predict the next $\tau$ timesteps for the complete test set. Using this counterfactual test set allows us to test how well a model generates counterfactual forecasts, which is impossible with real data. Furthermore, we can compose different datasets (both training as well as testing) with different $\zeta$, increasing or decreasing the influence of a time-dependent confounding factor or a treatment factor in the data.

\section{HG-t architecture}
As we have in Table~\ref{tab:benchmarks}, we provide an architectural picture for HG-t--- an extension of \citet{hassanpour2020} to the temporal setting ---in Figure~\ref{fig:HGt}. We have also included the architectural picture of DCRN, for comparison. The major take-away from Figure~\ref{fig:HGt} is the location of the recurrent structures. In DCRN, each factor is built from the same recurrent model, while in HG-t, each factor has it's own recurrent model. The latter does not allow for factors influencing each other over time, while DCRN does allow for this.

\end{document}